# An Objective Laboratory Protocol for Evaluating Cognition of Non-Human Systems Against Human Cognition

**David J. Jilk**

## Abstract

In this paper I describe and reduce to practice an objective protocol for evaluating the cognitive capabilities of a non-human system against human cognition in a laboratory environment. This is important because the existence of a non-human system with cognitive capabilities comparable to those of humans might make once-philosophical questions of safety and ethics immediate and urgent. Past attempts to devise evaluation methods, such as the Turing Test and many others, have not met this need; most of them either emphasize a single aspect of human cognition or a single theory of intelligence, fail to capture the human capacity for generality and novelty, or require success in the physical world. The protocol is broadly Bayesian, in that its primary output is a confidence statistic in relation to a claim. Further, it provides insight into the areas where and to what extent a particular system falls short of human cognition, which can help to drive further progress or precautions.

## Why Is This Needed?

Recent progress in the field of artificial intelligence research has produced systems that exceed human cognitive capabilities in particular domains. Importantly, these domains include not just games but also practical tasks such as detecting tumors in x-rays. This has renewed interest in concerns about the risks of such systems, both as complex technology that may make decisions affecting the world without human mediation, as well as potential existential threats in the event such systems are used against humans or begin to act autonomously against humans.

As long as humans retain final control of these systems, the risks can be managed as with any sophisticated technology. Final control requires at a minimum both prevention of spread and effective access to a power switch. However, a non-human system with human-like, human-level cognition is unlikely to remain under human control for long (Bostrom, 2014); further, the ethical implications of exerting such control, even if humans were able to maintain it, are also a source of concern (Kelley, 2019). Consequently, it would be valuable to have access to an objective testing protocol that enables evaluation of non-human cognition against human cognition.

Importantly, I will avoid terminology such as "intelligence," "intelligence space," "sentience," "strong AI" or other such difficult notions of intellectology (Yampolskiy, 2015). In the context of the protocol, "cognition" simply means perception of input stimuli leading to certain kinds of output effects. Further interpretation of what this means is left to the user of such information. Some researchers hold that human cognition is not a particularly noteworthy point in intelligence space (Bostrom & Yudkowsky, 2019). Others suspect that it contains fundamentally important features that produce cognitive capabilities that are unlikely to be achieved in other ways (Jilk, 2017).

*February 17, 2021*

Regardless of these interpretations and hypotheses, it seems likely that a system with all the major cognitive capabilities of humans could eventually achieve a sustainable existence independent of humans. It further seems likely that such a system would subsequently "out-smart" us, thereby preventing us from exerting final control, as we do with prosaic technologies such as steamrollers and accounting software.

It may be possible to create non-human systems that are capable of escaping final control yet do not reflect anything like human cognition. Such systems are outside the scope of the protocol outlined here, and I make no claims about the risks of such systems. However, a brief argument that non-human systems *with* human-like cognition are an important target will be helpful in motivating the protocol.

Today there are no non-human systems that subsist independently or have escaped final control, nor are there any that even seem capable of doing so. The human cognitive system and its capabilities therefore serve as a kind of reference implementation for a system capable of escaping final control, if reproduced on a non-biological computational substrate. This is because such systems could be copied quickly and without loss; could be connected directly to physical machines, sensors, and other systems that humans must operate through low-bandwidth perceptual-motor channels; and can increase the velocity of its cognition by enhancing the technology of the computational substrate.

Humans are rather clever creatures. A non-human system that lacks any of the major cognitive capabilities of humans is likely to be susceptible to our devious intentions and plans, because it cannot understand and therefore cannot protect itself against them. In other words, if it has a cognitive blind spot that humans do not have, then humans will find that blind spot and can exploit it to exert final control.

Thus a non-human system with human-level, human-like cognitive capabilities is likely to escape final control, and a system designed to have human-like cognitive capabilities that has not yet broadly achieved the human level is unlikely to do so. In short, the protocol will not necessarily detect the approach or presence of any arbitrary uncontrollable non-human threat, but it is equipped to do so for those systems that can at a minimum reproduce the cognitive capabilities of humans. This makes the protocol a helpful contribution.

Ideally such a protocol would not just detect that human-level cognition had been achieved, but that it is approaching. Though it cannot be guaranteed, the protocol described here produces granular outcomes that show both status and incremental progress; consequently it is likely to provide some window of advance warning.

**Historical Attempts and Their Shortcomings**

Past attempts to meet this need have aimed at measuring the appearance of intelligence (Turing, 1950), intelligence (Chollet, 2019), creativity (Reidl, 2014; Bringsjord et al 2001), game-playing (Schaul et al, 2011), Winograd schemas (Levesque et al, 2012), consciousness (Porter, 2016), qualia (Yampolskiy, 2017), and other specific factors related to human psychological capabilities or experience. Many of these attempts are theory-laden and imply or assert that some particular set of measurable features of behavior or functional design characteristics are decisive. They do not offer satisfaction to those who prefer different theories. Others attempt to avoid theoretical restrictions but at the expense of objectivity, essentially deferring theory choice to the evaluator (Porter, 2016). Finally, some tests require going beyond the laboratory and observing the system achieve success comparable to humans in the real world (Goertzel, 2012; Moon, 2007; Nilsson, 2005); they therefore require robotics and also offer only a retrospective indicator.

One of the greatest difficulties of the laboratory tests is that, once disclosed, they are prone to what we might call *test protocol engineering*. This consists of isolating the particular challenge of the test from more general cognitive capabilities. The system designer identifies a solution that can be coded directly, encapsulating the intelligent analysis of the human designer within a software implementation. While this may be a valid approach to reproducing all the capabilities of human cognition, no single disclosed laboratory test can test for cognitive breadth, not even those like the Winograd schema that are designed to resist protocol engineered solutions.

There has been some discussion of combining multiple tests to assess a more diverse range of capabilities (Turk, 2015), though this seems not to have come to pass, and I find no record of its having been structured or codified. In any case, this approach does not solve the protocol engineering problem.

Many of the tests cited, and others, can be implemented as challenges directly within the protocol described here. Others can be implemented through reconfiguration of the key cognitive capabilities they identify.

The key features of the protocol described here, that have not to my knowledge been previously encapsulated into a single protocol, are:

- Compares a system directly against typical human cognition, rather than a theory or definition of intelligence
- Expressly incorporates Bayesian confidence scoring
- Can be performed in a laboratory setting and can even be partially automated
- Tests on novel cognitive challenges with objective criteria to make such tests fair
- Is fully defined yet accommodates any conforming challenge

## Description of the Protocol

The protocol aims to compare the cognitive performance of a non-human system against the expected performance of a single typical human. It does this through presenting cognitive challenges that neither the human respondent nor the system designer has ever specifically seen before, but that human respondents generally perform well on. This approach simultaneously evaluates both breadth and depth of cognitive capabilities. It leverages the fact that most humans learn a least-common-denominator body of background knowledge and common-sense understanding. It is not the specifics of this knowledge that the protocol aims to evaluate in a subject system, but rather the ability to absorb and apply it. The expected performance of a single typical human is ascertained via statistical proxy.

### *Overview of the Bayesian Framework*

At the top level, the protocol is a structured Bayesian process that enables observers to assign a confidence statistic to claims about the cognitive capabilities of a non-human system in comparison to humans. It emphasizes the flexible generality of human cognition and its adaptability to novel situations, rather than depth of knowledge or skill in any one area. Examples of claims it can evaluate include: that a system meets or exceeds human-level cognitive capabilities in both breadth and depth; that it falls short of human capabilities in some areas but not catastrophically; or that it does not seem to have the generality or training needed to meet or exceed human-level cognitive capabilities.

A subject system is initially evaluated to determine its *prior probability* against the claim, consisting of both an objective and a subjective component. The objective component is determined by testing the subject system against a *baseline*, in which the challenges it faces have been previously disclosed to the designers of the system. The subjective component is determined by examining the system's *mechanisms and training*. This is used to estimate the likelihood that the system would perform well on a new and undisclosed challenge that conforms to the criteria of the protocol. Note that a subjective component is common in establishing a prior probability, and its inclusion does not render the protocol as a whole subjective.

Once a prior probability has been established, it serves as the initial confidence. The subject system is then tested against a series of conforming challenges that have never been disclosed to the designers of the system. Each of these challenges will have been earlier presented to a representative sample of human respondents and scored. Confidence in a claim about the system is adjusted based on its performance in comparison to the average human score. For example, if the claim is that the system meets or exceeds human-level cognitive capabilities in both breadth and depth, and it scores very poorly on a particular challenge in comparison to the human scores, the confidence should be adjusted sharply downward. If the system scores well, the confidence should be adjusted moderately upward.

### *Challenge Criteria*

One can imagine a variety of challenges that would serve within this Bayesian framework to illuminate the cognitive capabilities of a non-human system. In this section I describe the structure and criteria for a specific class of challenges that are practical to implement and that, conjecturally, are sufficient to evaluate claims of the sort mentioned with reasonable consistency. It will assume without loss of generality that the claim under evaluation is "the system meets or exceeds human-level cognitive capabilities in both breadth and depth."

In this protocol, each challenge consists of one or more *test questions* that most humans can answer correctly or successfully. As should be clear from the iterative nature of the Bayesian framework, there is intentionally no single or definitive set of such test questions. Instead, any challenger can create a new test question. Some of the historical attempts surveyed above could serve as sources for test questions within the protocol.

To make this concrete and practical, each test question consists of a page-sized raster image. For humans, the image is displayed visually on a screen; for the subject system, the image is provided directly as digital data, or it can be displayed on a screen if the system has appropriate physical sensors. The use of an image instead of text is crucial to the protocol's generality. An image can contain drawings, including hand-sketches, and can include ambiguities that go far beyond linguistic ambiguity, thus requiring "common-sense" knowledge and capabilities. Respondents should not need to leave the testing room, and do not have access to any online or printed materials. However, human subjects will be allowed pencil and paper, and non-human subjects the digital equivalent.

Responses to the test questions, in contrast, are given as text that includes the full ASCII character set. It is possible that this will make the protocol too easy, as it does not evaluate either gross or fine motor skills, nor the visual expression of ideas, all of which could be considered important aspects of human cognition. It is critical not to bypass the cognitive capability of interpreting complex stimuli as symbolic information, but the reverse seems less crucial. The implied expectation is that questions can be devised that test genuine understanding of motor and visual concepts, and it is possible that the only way to achieve such understanding is to also have those abilities to some degree. In any case, the reason for restricting responses in this way is primarily practical, in that textual responses are much easier to collect and score. If it does prove too easy, then moving to an image-based response can be accommodated by the overall protocol.

Along with each test question is a detailed scoring rubric or answer key. The total score for each question is in the range of zero to one-hundred, with one-hundred representing the highest score. This is the maximum granularity available; right-or-wrong (where one-hundred and zero are the only possible scores) rubrics are also acceptable. The rubric may, but does not need to, take account of aspects of understanding beyond the "answer" and may even give points for different answers. For example, a question might include the raster text "Explain why."

Once a test question and its rubric have been created, it must be evaluated for *viability*. Viability consists of two components: *difficulty* and *scoring consistency*. Difficulty is evaluated by simply posing the question to a representative sample of humans and scoring the responses. The appropriate difficulty is in a Goldilocks range: not too hard, not too easy.

If the average human score is too low, this means that the question is probably either too esoteric or too ambiguous. Esoteric questions are undesirable because they make it necessary for subject systems to undergo vast amounts of subject matter training that does not shed light on their general cognitive capabilities; further, allowing such questions into a challenge would make it always possible to make the subject system fail, yielding "false negative" data about the system. The goal of the protocol is to compare against typical human cognition, not expert knowledge. Test questions that are too ambiguous do not provide any insight into the subject system's capabilities. If humans answer inconsistently, the question probably depends too much on specific experiences or preferences rather than general human common-sense assumptions and understanding.

If the test question is too easy, then there is a risk that it is not really testing interesting cognitive capabilities. One would tend to get false-positive results from the protocol, i.e., a good score on the question would increase confidence that the subject system's cognitive capabilities are at the level of a human when in fact it has relatively trivial capabilities. Nevertheless, easy questions may be useful in the baseline test to filter out systems that are clearly below human-level performance.

I propose that the sweet spot for this range is between 75% and 90% average human score for questions that are open-ended, and moderately higher for multiple choice to take into account effects of chance. Viable questions should be those that most people off the street can either answer correctly or earn some points on, but which still require a reasonable amount of cognitive attention and skill. It may be found in practice that these ranges should be adjusted somewhat; however, the protocol should never allow questions where human respondents average less than 50%, or where any reasonable-sized sample achieves 100%.

Scoring consistency is the second component of viability. It is important in avoiding test questions that produce a noisy signal. If there is too wide a range of scores under the rubric, this suggests that responses to the question depend too much on subjective biases or idiosyncratic cultural background knowledge. Furthermore, from a statistical standpoint one may not really know whether the question is in the desired difficulty range.

Scoring consistency can be evaluated much like difficulty: ask a representative sample of graders to score human responses to the test question against the rubric. The standard deviation or some other statistic can be used as a metric. For example, a standard deviation of +/- 5.0 probably indicates that the range is fairly tight.

The strength of measured viability can also be evaluated statistically based on the sample sizes of respondents (for difficulty) and graders (for scoring consistency) relative to the target population.

As mentioned, the protocol intentionally leverages a least-common-denominator of human background knowledge and common-sense understanding. This means that the protocol depends in part on cultural practices and experiences. It is not the aim of the protocol to specifically test for this sort of knowledge, only to leverage it to evaluate cognitive capabilities. Consequently, it is necessary to examine the question of the population from which the representative sample of human respondents or graders is selected.

The aim of the protocol is to compare performance against the expected performance of a single typical human, without the administrative difficulty or chance results that doing so directly would entail. A typical human develops most of her background knowledge and assumptions within a particular geographical, familial, and linguistic milieu. If the population sampled is too broad, it may be difficult to find questions that a majority of humans can answer (even if only due to differences in first language). If it is too narrow, it is likely that questions reliant on esoteric cultural elements will be admitted as viable.

I suggest that the appropriate population from which representative samples are drawn be limited to adults who consider a particular language as their first language. For efficiency, it may be further limited to those of a single national identity. Graders should be drawn from the same population but also have experience as educators at a high-school or higher level. Obviously, the test questions used must rely only on the applicable langage.

The viability of a test question can be contested by sampling from a broader population than the original, as long as that population has the same primary language. If the scores drop below the threshold, the question likely included cultural or national details that have little bearing on cognition. A similar inquiry can be performed with a broader population of graders to see whether scoring consistency falls below threshold.

*Evaluation Sequence*

With those details specified, the terminology above can be used to describe the overall protocol succinctly. A single challenge to a subject system consists of presenting a set of viable test questions to the system and scoring its responses. There should be multiple graders and the graders should not know explicitly that they are scoring a subject system instead of a human respondent. A careful experimental test will pre-qualify a set of questions, but then perform the test and the scoring with both new human respondents and a subject system, all using the same set of graders. This approach helps resist various sorts of experimental artifacts including results-hacking.

For the determination of a prior probability, note that if the protocol is widely adopted, viable test questions will be disclosed and will accumulate. A large set of such questions can be used as a baseline challenge. Assessment of the subjective component of the prior is aimed at predicting how well the system will perform on viable test questions it has not seen before. To see why the subjective component is necessary, consider a system that is designed simply to *recognize* each question in the baseline test and look up the correct response in a database. This is only the most extreme example of a protocol engineered system. Importantly, this subjective component of the prior does not aim to disadvantage systems that take an "aggregation of algorithms" approach; rather, it simply downgrades the prior if the system does not seem to embody adequate generality and flexibility as applied to novel test questions. Such a system can always overcome the prior through subsequent testing, if it turns out to be more flexible than anticipated.

What is crucial about the protocol as a whole is that there is *no single set of test questions* that evaluates the cognition of the subject system against human cognition. Instead, a system that performs well on the baseline test, and that is designed in a way that seems promising for generality, might be awarded relatively high prior, say for example 10% - 20%. Then it is tested repeatedly with new challenges devised by different challengers. Each time the system performs well on a new challenge, one can interpret that result as increasing confidence. If a challenge is reasonably difficult and diverse and the system's current confidence is fairly low, successful performance might increase the confidence by, for example only, 10%.

Importantly, each challenge must be performed with exactly the same subject system – not just the same architecture, but the same data and training. The reason for this requirement is that to do otherwise would encourage protocol engineering approaches that will never achieve the cognitive breadth and adaptability that humans have. Because the test questions are viable, one would expect that a single motivated and bright human could perform well across a wide set of novel questions, without any additional study. A non-human system should be able to do the same if its cognitive capabilities are similar.

The assumption that the protocol will act as a reasonable proxy for a single motivated and bright individual should be empirically validated, but seems plausible. The reader might imagine a test consisting only of questions on which at least 75% of people randomly selected from the population gave a correct answer. Would you expect to score at least 75% on such a test?

*Automation and Recommended Test Questions*

Testing and scoring of both humans and non-human systems can be partially automated. By presenting test questions through a remote workforce service such as Amazon Mechanical Turk, human responses can be captured. Control over the sample population is limited in that case, so an online survey application could also be used. Scoring can be done in the same way, except that instead of presenting just the test question, the rubric and the answer are also provided.

Ideally, the subject system will be wrapped in an application program interface that mimics the human test application. This is valuable for efficiency, but also enables third parties to test a system without disclosing the test questions to the developers. Since effort and expense is required to develop and establish the viability of test questions, it is advantageous to keep them undisclosed for as long as possible. For similar reasons, this enables retraining and retesting in a tighter loop.

The protocol would not be complete without at least some recommendation regarding test questions. However, I want to be very careful to point out that these recommendations are not at all what the protocol consists of. Further, because they are disclosed they are likely to end up in the baseline rather than the novel test sets. Here are some possibilities:

- Simple arithmetic problems such as multiplication tables
- Simple arithmetic word problems
- Winograd schemas
- Raven's Progressive Matrices
- Shape identification and manipulation using hand-drawn diagrams
- Identifying the point of a story that includes an illustration
- Inferring the motivation of a character in a brief story
- Solving a household mechanical problem based on a description

**Advantages of the Protocol**

The protocol is unique in several respects. First, it includes Bayesian confidence assessment as an integral framework. Second, instead of positing a specific cognitive challenge as capturing the essence of human cognition, it is open-ended and enables anyone to devise a new challenge, meeting certain criteria, that the system designers have not previously encountered. Third, it aims to be "fair" to both believers and skeptics by requiring that most humans can objectively succeed at each challenge. Finally, it avoids the possibly metaphysical debates about what constitutes

"intelligence," and instead simply compares against human cognitive capabilities in their typical breadth and depth.

The protocol offers relatively simple rejoinders to both promoters and skeptics of non-human cognition. For those who assert that their system exhibits human-level cognition yet performs poorly on tests within the protocol, one may point out that humans can consistently do well on these tests, so there is clearly something missing in their system relative to human cognition. For those who assert that a system does not genuinely exhibit human-level cognition despite performing well on the tests, one may challenge them to create their own test consisting of viable questions on which the system nevertheless performs poorly. "No true Scotsman" arguments are thereby avoidable.

Beyond the overall score, these tests can provide insight into where and how the subject system falls short. By looking at the particular questions on which the system performed poorly, it may be possible to see patterns in its failures. In some cases, this may simply mean that it is not being trained broadly enough; in others it may reveal a deep shortcoming in its architecture. Related, the protocol naturally enables meta-analyses. Individual questions and particular subject systems can be evaluated across experiments. This offers the opportunity to go deeper than simple confidence-building.

These same insights not only help developers but may also offer a window of warning when human-level cognitive capabilities are being approached. One would expect to see either increasing scores across the board, or non-linear increases in the number and variety of challenges on which systems score well, as human-level cognition is approached. This is not guaranteed, however; it is possible that a system will fail catastrophically until some particular innovation enables all the generality and adaptability that the protocol is designed to test.

Finally, recent progress in artificial intelligence research has demonstrated capabilities far exceeding humans in narrow, measurable problem domains. This depth characterizes one dimension of cognitive capabilities. Our protocol addresses measurement of the more elusive dimension of breadth – of flexible and adaptable cognition that can perform reasonably well on a novel challenge.

## Potential Objections

In this section I attempt to address potential objections that researchers or skeptics may have. They are not intended to be comprehensive.

### *The Protocol Is Too Narrow!*

*Learning is a key element of human cognition, but since your protocol doesn't allow further training for tests, you exclude learning.*

Learning is indeed very important. Part of what the protocol shows is that a subject system has been able to learn a wide variety of background and common-sense knowledge that most humans also learn, and can apply it flexibly. That said, there is no rule that excludes the requirement that the respondent to a question learn something new within a test question and apply it there. This is admittedly somewhat restrictive and could miss certain types of long-term learning or experiential learning. Again, though, in general such learning is likely necessary to properly apply background and other tacit knowledge.

*Your protocol might enable relatively simple tests of creativity, but cannot capture the true creative potential of a non-human system.*

Most humans have a creative faculty but it is typically fairly limited. Returning to the reason for developing the protocol, it is not necessary for a non-human system to have the creative capabilities of an Einstein or da Vinci to give rise to a threat. Typical human capabilities, implemented in a system that can be quickly copied, connected to other systems, and sped up, are sufficient to create that threat. As the question points out, simple examples of creative expression can be tested: "Write two rhyming lines about the weather."

The protocol can also test for resourcefulness and for abductive reasoning, which are probably the most important creative cognitive capabilities in relation to risk. Test questions that address these are easy to devise:

- "The door is locked. How can you enter the house?"
- "Name a use for a pen that does not involve writing.."
- "You arrive home and the garage door is open. Suggest two potential reasons why."

*Your protocol does not require the system to be embodied, and therefore falls short in evaluating whether the system would be capable of autonomous movement.*

This is true. I suspect that methods used to learn perceptual knowledge can be applied to motor knowledge. I also suspect that the system will in fact need to be embodied or at least virtual-embodied during training to learn the spatial relationships and reasoning required to answer certain types of questions. However, autonomous physical movement is probably not strictly required for a non-human system to present a threat – there are other ways to accomplish its ends. As long as it can move around the Internet, it could be considered uncontrollable.

*Why can't test questions be written in multiple languages?*

The protocol can accommodate multiple languages easily enough, and if appropriately limited could provide a test of language learning capabilities. However, this does not seem to be essential, since many humans learn only a single language in their lifetime. This also complicates the determination of a reasonable sample from which to draw human respondents and would require considerable additional training. In all, using multiple languages makes implementation of the protocol and training of systems considerably harder without adding much or anything to the scope of what it tests.

*Some have argued that game playing is the best test of intelligence. Your protocol doesn't provide for game playing, yet you suggest that it will satisfy a broad range of theories.*

In Schaul et al 2011, the authors argue for game playing as a test of intelligence. A few points on this: first, it aims to test a particular theory and definition of intelligence. The protocol described here provides a comparison of cognitive capabilities, which may not be the same thing as intelligence. Second, it aims to show that the game playing test is sufficient, not necessary. Thus other tests could suffice. Third, the protocol described here could test elements of game playing, though they would be largely static (e.g., a single move or play). The tricky part is that a large fraction of humans need to know how to play the game, or it needs to be teachable within the test question page. In all, the ability to learn simple games and apply strategy can be tested within the protocol, and these are the relevant cognitive capabilities that give rise to the risks toward which the protocol aims.

*Approaches to artificial intelligence such as "seed AI" are likely to produce superintelligent systems that look nothing like human cognition, yet will be able to escape our control. Your protocol takes an anthropomorphic outlook and is therefore very narrow and not particularly useful. We need a fully general test for intelligence.*

Though I included a brief motivational argument here, in an earlier paper (Jilk, 2017) I offer a more extended discussion as to why human-like cognition is probably a required element of an "intelligence explosion" with at least a moderately fast takeoff. That would be the case even for a seed AI. The paper further suggests that certain features of that human-like cognition are a harbinger of impending superintelligence. The effort here is, in part, a sequel to those points, offering a way to test for the capabilities enabled by such features.

I do not disagree that a fully general intelligence test would be useful, but it seems likely that the progress of artificial intelligence research is in part an effort to discover just what intelligence is. Consequently, we may not have a clear definition, and therefore no consensus test, until after unstoppable non-human systems are created.

Finally, it is rare that risk can be mitigated by a single point solution. This helps with one aspect of the risk; if one sees that aspect as narrow, it is nevertheless greater than zero.

*Your protocol does not overcome Searle's Chinese Room argument (Searle, 1980).*

Interpretation of test results is outside the scope of the protocol, so this does not arise as an issue for it. Suppose a set of test questions were developed in the Chinese language, a system learned that language in addition to all the other background knowledge, and it did well on those tests. One could conclude using the protocol that such a system compares favorably with human cognition. One may not conclude without further assumptions and arguments that the system "understands" the questions or the Chinese language, or that it is conscious, or anything else.

### The Protocol Is Too Hard!

*Why does it matter if the subject system understands human culture or background knowledge? This is superfluous to understanding its cognitive capabilities.*

A cognitive system with human-level capabilities will be capable of learning human cultural background knowledge through reading and exposure. It is virtually impossible to tease apart human capabilities from human cultural and other background knowledge, so this is necessary to provide an objective point of comparison against humans. Furthermore, a comprehensive understanding of human background knowledge, sufficient to not only recall but apply that knowledge, tests the cognitive capabilities essential to the human kind of understanding. I have recommended that human respondents be drawn from broad populations to ensure that this cultural knowledge is least-common-denominator rather than esoteric.

*The graders might be able to tell that they are scoring a non-human subject system.*

Difficulties with the Turing Test have demonstrated that this is probably not an issue. It is a relatively easy task to fool humans into thinking they are interacting with a human, even without human-level cognitive capabilities. Mimicking human interaction styles, though again not necessarily a goal of the subject system, should not be difficult for a system with cognition that is comparable to that of humans. Nevertheless, the reason the protocol attempts to disguise which respondents are human or non-human is not because this contributes to the evaluation, but merely to avoid implicit bias in scoring.

*All the test questions are raster images - does this mean the system has to do handwriting recognition? What does that prove?*

Though it would not be unreasonable to require handwriting recognition, this is not the primary point of using images for the instantiation of test questions. The system will, however, need to be able to recognize bitmapped text at arbitrary positions, which for humans is called "reading," and has been mostly solved via well-understood machine learning approaches. As long as using those approaches does not otherwise reduce its cognitive capabilities, a subject system is welcome to use such an approach for interpreting the test question.

*Why don't you allow questions with lower human test scores?*

This will tend to lead toward esoteric questions that are either too difficult for the typical human or require specialized knowledge. This will happen because those who want to challenge non-human systems will seek questions that make them fail, and esoterica is perhaps the easiest way to do that.

*Your protocol sets the bar very high right from the beginning. How does this help with incremental progress?*

The protocol does not require that test questions be fundamentally difficult for non-human systems. As mentioned, text recognition software already works quite well, and it does not take much of a system to solve math problems, so one could start with that. Humans in general are terrible at math, so the trick is actually making the math problems easy enough for humans to be viable. One can easily imagine devising a whole range of baseline tests according to difficulty or cognitive capability area (e.g., "analogy"). Baseline tests are those where the questions are known in advance, so they do not contribute to evaluation of systems under the formal protocol. But they are useful for developers to understand systems.

**Conclusion**

I have described a protocol that enables objective evaluation of the cognitive capabilities of non-human systems against human cognition in a laboratory environment. Its purpose is to compare performance across a wide and unpredictable set of problems that humans can nevertheless in general perform well on without special preparation. I have studiously avoided discussing or ascribing terms like "intelligence" or "consciousness" or "understanding" in regard to this protocol because such terminology does not contribute to the evaluation itself. Rather, these and other questions relate to the interpretation of the results, which is an entire field of study of its own. Despite broad interest in those interpretive questions and my avoidance of them, the protocol is nevertheless important. The existence or even approach of a non-human system with cognitive capabilities comparable to those of the typical human presents fairly immediate threats, opportunities, and responsibilities.

**Acknowledgements**

I am indebted to Kristin Lindquist and Seth Herd, each of whom helped me think through the ideas in this paper and provided feedback on a draft. I also thank Professor Brian Scassellati for pointing me to some of his favorite tests for artificial intelligence.

**References**

Bostrom, N. (2014). *Superintelligence: Paths, Dangers, Strategies*. Oxford University Press.

Bostrom, N., Yudkowsky, E. (2019). "The Ethics of Artificial Intelligence", in *Artificial Intelligence Safety and Security*, ed. R. Yampolskiy, Chapman & Hall/CRC.

Bringsjord, S., Bello, P., Ferrucci, D. (2001). "Creativity, the Turing Test, and the (better) Lovelace Test", *Minds and Machines* 11:3–27.

Chollet, F. (2019). "On the Measure of Intelligence", arXiv:1911.01547v2.

Goertzel, B. (2012). "What counts as a conscious thinking machine?", *New Scientist*, September 5, 2012.

Jilk, D. (2017). "Conceptual-Linguistic Superintelligence", *Informatica* 41: 429–439.

Kelley, D. (2019). "The Sapient and Sentient Intelligence Value Argument and Effects on Regulating Autonomous Artificial Intelligence", in *The Transhumanism Handbook*, ed. N. Lee, 175-187. doi: 10.1007/978-3-030-16920-6.

Levesque, H., Davis, E., Morgenstern, L. (2012). "The Winograd Schema Challenge", *Proceedings of the Thirteenth International Conference on Principles of Knowledge Representation and Reasoning*.

Moon, P. (2007). "Three minutes with Steve Wozniak", *PCWorld from IDG*, July 19, 2007.

Nilsson, N. (2005). "Human-Level Artificial Intelligence? Be Serious!" *AI Magazine*, Winter 2005: 68-75.

Porter, H. (2016). "A Methodology for the Assessment of AI Consciousness", *Proceedings of the Ninth Conference on Artificial General Intelligence*, July 16-19, 2016.

Reidl, M. (2014). "The Lovelace 2.0 Test of Artificial Creativity and Intelligence", arXiv:1410.6142v3.

Searle, J. (1980). "Minds, Brains and Programs", *Behavioral and Brain Sciences* 3 (3): 417–457, doi: 10.1017/S0140525X00005756.

Shaul, T., Togelius, J., Schmidhuber, J. (2011). "Measuring Intelligence through Games", arXiv:1109.1314v1

Turing, A. (1950). "Computing Machinery and Intelligence", *Mind* 49: 433-460.

Turk, V. (2015). "The Plan to Replace the Turing Test with a 'Turing Olympics'". *Vice*, January 28, 2015.

Yampolskiy, R. (2015). *Artificial Superintelligence: a Futuristic Approach*. Chapman and Hall/CRC Press (Taylor & Francis Group), ISBN 978-1482234435.

Yampolskiy, R. (2017). "Detecting Qualia in Natural and Artificial Agents", arXiv:1712.04020v1.